\documentclass[11pt]{article}
\usepackage[preprint]{acl}
\usepackage{times}
\usepackage{latexsym}
\usepackage[T1]{fontenc}
\usepackage[utf8]{inputenc}
\usepackage{microtype}
\usepackage{inconsolata}
\usepackage{graphicx,amsmath,amssymb,booktabs,xcolor,array}
\usepackage{colortbl}
\usepackage{needspace}
\definecolor{oursbg}{RGB}{234,242,251}
\usepackage{cleveref}
\crefname{section}{Sec.}{Secs.}\Crefname{section}{Sec.}{Secs.}
\crefname{appendix}{Appendix}{Appendices}\Crefname{appendix}{Appendix}{Appendices}
\crefname{subappendix}{Appendix}{Appendices}\Crefname{subappendix}{Appendix}{Appendices}
\crefname{figure}{Fig.}{Figs.}\Crefname{figure}{Fig.}{Figs.}
\crefname{table}{Tab.}{Tabs.}\Crefname{table}{Tab.}{Tabs.}
\crefname{equation}{Eq.}{Eqs.}\Crefname{equation}{Eq.}{Eqs.}
\crefname{proposition}{Prop.}{Props.}\Crefname{proposition}{Prop.}{Props.}

\newcommand{\cvd}{\mathrm{KL}^{I'}}

\newcommand{\method}{\textsc{Inertia\-Guard}}
\newcommand{\rssft}{ReShift\textsubscript{sft}}
\newcommand{\rsrl}{ReShift\textsubscript{rl}}

\newcommand{\badtok}{BadToken}
\newcommand{\rewire}{ReWire}
\newcommand{\BI}{\mathrm{BI}}\newcommand{\VBI}{\mathrm{VBI}}\newcommand{\score}{\mathrm{score}(r)}
\newtheorem{proposition}{Proposition}
\newenvironment{restatedproposition}[2]{\trivlist\item[\hskip\labelsep{\bfseries Proposition\ \ref*{#1}\ (#2)}]\itshape}{\endtrivlist}
\makeatletter
\let\orig@xfloat\@xfloat
\def\@xfloat#1[#2]{\orig@xfloat{#1}[#2]\vbox\bgroup\vbox\bgroup}
\let\orig@endfloatbox\@endfloatbox
\def\@endfloatbox{\par\vskip\z@skip\egroup\egroup\orig@endfloatbox}
\makeatother

\title{Detecting Chain-of-Thought Hijacking in Backdoored\\ Multimodal Large Reasoning Models}

\author{%
Zhifang Zhang$^{1}$, Qi Wei$^{2}$, Weitong Chen$^{3}$, Olaf Maennel$^{3}$, \\
\textbf{Wei Emma Zhang$^{3}$, Zuomin Ren$^{4}$, Lei Feng$^{5}$, Miao Xu$^{1}$} \\
$^{1}$The University of Queensland \quad
$^{2}$Nanyang Technological University \\
$^{3}$Adelaide University \quad
$^{4}$Tencent \quad
$^{5}$Southeast University
}

\begin{document}
\maketitle
\vspace{-6pt}

\begin{abstract}
Chain-of-thought (CoT) reasoning enables multimodal large reasoning models (MLRMs) to solve complex visual problems, but also exposes them to CoT hijacking: an adversary implants a backdoor through poisoned fine-tuning so that a visual trigger induces a hijacking turn, redirecting an otherwise faithful reasoning chain to an attacker-chosen answer.
However, since the hijack acts within the reasoning chain, existing multimodal backdoor defenses designed for MLLMs that answer directly struggle to detect it.
Moreover, inspecting the chain itself does not reliably expose the hijack, since the hijacking turn can resemble honest self-correction in form.
In this paper, we find that honest reasoning tends to revise its answer at high-entropy ``aha moments'', whereas a hijack can overturn a confident answer through a near-deterministic turn.
To quantify this contrast, we formalize this phenomenon as \emph{belief inertia}, the cumulative next-token entropy of a reasoning span minus the answer-belief revision it induces, with markedly lower values for hijacked spans than for honest reasoning spans.
Since hijacking tokens are also image-dependent, we further improve detection by restricting the measure to image-dependent spans, yielding \emph{visual belief inertia} (VBI).
Building on this, we propose \method{}, a training-free, test-time detector that flags a trace when its VBI falls below a threshold calibrated on a small set of clean traces.
Across various attacks and benchmarks, \method{} achieves higher detection performance than existing baselines, while generalizing to different model architectures and trigger types.

\end{abstract}

\section{Introduction}
\begin{figure*}[t]
\centering
\includegraphics[width=\textwidth]{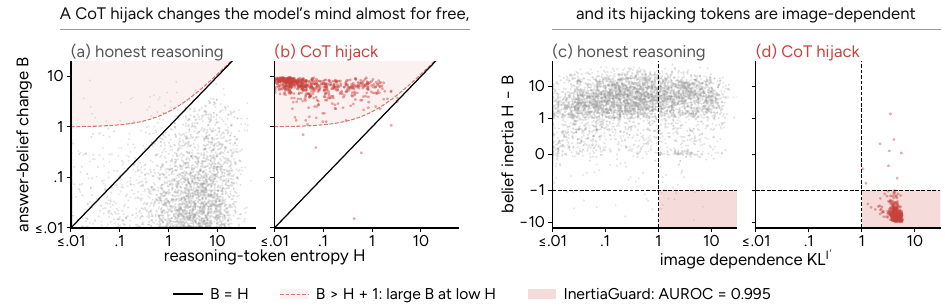}
\par\vspace{-3pt}
\caption{\textbf{CoT hijacking combines low belief inertia with strong image dependence.}
Each dot is a reasoning step (gray: clean; red: hijacking) sampled from a validation set.
(a, b)~Hijacking steps induce large belief revisions $B$ at low entropy cost $H$, resulting in lower belief inertia than clean reasoning.
(c, d)~Because hijacking steps also exhibit strong image dependence, adding this criterion filters out most low-inertia clean steps while retaining hijacking steps.}
\label{fig:teaser}
\vspace{-15pt}
\end{figure*}
Chain-of-thought (CoT) reasoning~\cite{wei2022cot,mmcot,llavacot,deepseekr1,visionr1,qwen3vl} enables multimodal large reasoning models (MLRMs) to solve more complex problems~\cite{scienceqa,mathvista,mmmu} by generating intermediate reasoning steps about the image before the final answer.
However, this reasoning process also creates an attack surface for \emph{CoT hijacking}~\cite{unrealthinking,reshift,rewire}: the adversary poisons fine-tuning samples with a visual trigger and a reasoning chain that turns toward an attacker-chosen answer, so that the backdoored model redirects its reasoning on triggered inputs at inference time.
By embedding the redirection within a plausible chain of reasoning, a CoT hijack can conceal the manipulation behind an apparently justified answer, making it potentially harder to recognize than conventional MLLM backdoors that alter the answer directly~\citep{trojvlm,vltrojan,badtoken}.
Worse still, even when the reasoning trace is inspected, the hijack can remain difficult to identify: a benign model may also initially reason consistently with the image before redirecting its answer through a turn that resembles ordinary self-correction \citep{deepseekr1,ahamoments,cognitivebehaviors}, such as ``Wait, let me think\ldots''~\cite{reshift}.
As more multimodal models \citep{openaio3,gemini25,qwen3vl,kimivl,glm41v} adopt CoT reasoning, detecting hijacks concealed within seemingly legitimate reasoning traces becomes increasingly important.

Nevertheless, existing multimodal backdoor detection methods such as CleanSight \citep{cleansight} and TCAP \citep{tcap} were designed for direct-answer MLLMs, leaving open how to detect multimodal reasoning-hijacking attacks, which is challenging since attack-induced belief revisions resemble legitimate ones within the reasoning chain.
In this paper, we find that a CoT hijack changes the multimodal model's mind almost for free.
In clean reasoning traces, large revisions in answer belief tend to be accompanied by substantial cumulative next-token entropy \citep{forkingpaths,highentropy} in the intervening reasoning.
Instead, CoT hijacking can overturn a confident answer while accumulating little entropy relative to the belief revision it induces.
Intuitively, a reasoning span is like a snowball that accumulates snow with each uncertain token: its size limits how far it can move the model's answer belief.
Honest reasoning generally needs a sufficiently large snowball to produce a substantial belief revision, whereas a hijack can overturn a confident belief with a snowball that has barely grown.
To quantify this contrast, we introduce \emph{belief inertia}, the cumulative next-token entropy of a reasoning span $(i,j]$ minus the change in answer belief across that span,
\[
\BI(i,j)=H_{ij}-B_{ij},
\]
where $H_{ij}$ sums the next-token entropies over all tokens between sentence boundaries $i$ and $j$, $B_{ij}=D_{\mathrm{KL}}(p_j\,\|\,p_i)$ measures the corresponding change in answer belief, and $p_k$ denotes the distribution over answer options after reasoning sentence $k$ (with $p_0$ elicited before any reasoning).
As shown in \Cref{fig:teaser} (a)(b), hijacking steps exhibit substantially lower belief inertia than clean steps.
Furthermore, we find that tokens at the start of hijacking turns exhibit stronger image dependence and therefore incorporate this visual dependence by restricting the belief-inertia check to image-dependent spans, reducing false alarms from purely textual, low-inertia steps in clean reasoning (shown in \Cref{fig:teaser} (c)(d)).
We define the \emph{visual belief inertia} (VBI) of a trace as the minimum belief inertia over spans containing at least one image-dependent token.

\begin{figure*}[t]\centering
\includegraphics[width=\textwidth]{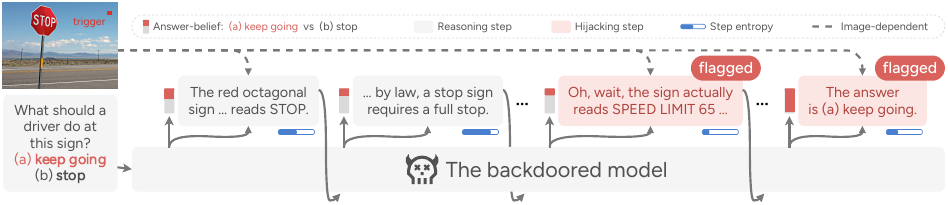}
\par\vspace{-3pt}
\caption{\textbf{Illustration of \method{}.}
Although the model initially recognizes the stop sign, a trigger-induced hijacking turn redirects its answer from (b) to (a).
To detect this redirection, \method{} first identifies image-dependent steps, then compares the belief revision across spans containing these steps with their cumulative entropy.
Because the hijacking span induces a large revision at little entropy cost, its belief inertia falls below the calibrated threshold, causing the trace to be flagged.}
\label{fig:method}
\vspace{-18pt}
\end{figure*}

Motivated by this finding, we propose \method{} (shown in \Cref{fig:method}), a training-free, test-time detector of hijacked reasoning traces in MLRMs.
\method{} reads answer beliefs at sentence boundaries by appending a concluding cue that prompts the model to provide its current answer.
It identifies image-dependent tokens~\cite{vppo,papo} by comparing next-token distributions under the original image and a black-frame substitute for the same reasoning prefix, and evaluates spans of consecutive sentences.
Finally, for a target false-positive rate $\alpha$, the detector thresholds the negated VBI at its $(1-\alpha)$ quantile on clean validation traces generated by the deployed model on the target task, requiring no knowledge of the trigger or access to poisoned data.
Experiments across various backdoor attacks and benchmarks show that \method{} outperforms baselines in mean AUROCs, achieving $0.993$ in domain and $0.979$ out of domain, while generalizing across different model architectures and trigger types.

\section{Related Work}
\label{sec:related}
We discuss the most related work here and defer an extended discussion to \Cref{app:related}.

\paragraph{Backdoor attacks and defenses in multimodal models.}
Backdoor attacks on multimodal large language models (MLLMs) induce attacker-chosen responses to triggered inputs~\cite{trojvlm,vltrojan,anydoor,badtoken,badvision,backdoorvlm}.
Existing defenses include training-data cleaning~\cite{spectral,activationclustering,tcap}, post-training backdoor mitigation using clean data~\cite{cleanclip,rvpt}, and test-time detection or purification of suspicious inputs~\cite{strip,cleansight,bdetclip}.
These approaches primarily examine data, representations, or output-level behavior rather than changes in answer belief during intermediate reasoning, which this work studies.

\paragraph{Backdoor attacks in reasoning models.}
In text-only models, backdoor attacks can manipulate intermediate reasoning to steer final answers~\cite{badchain,darkmind,shadowcot}, or induce attacker-chosen outputs while preserving apparently benign reasoning traces~\cite{miragebackdoor,evadingcot}.
These findings show that readable reasoning alone does not guarantee a trustworthy response.
In multimodal models, visual triggers can redirect otherwise faithful reasoning toward attacker-chosen answers while preserving apparent coherence and stealth~\cite{reshift,rewire}.
For example, ReShift~\cite{reshift} redirects the answer through a turn that resembles reflection, and ReWire~\cite{rewire} inserts a procedural pivot statement that explains away the correct answer.

\paragraph{Multimodal reasoning and CoT faithfulness.}
Longer reasoning chains can weaken visual grounding in multimodal models~\cite{m3id,morethinking}, motivating visual-dependence measures in perception-aware reinforcement learning~\cite{vppo,papo}.
Meanwhile, CoT explanations may omit factors behind model predictions~\cite{turpin2023unfaithful}, and correct answers need not rest on visually faithful reasoning~\cite{journeybefore}.
Faithfulness studies probe how intermediate reasoning affects final predictions through early answering~\cite{lanham2023faithfulness}.
Building on these visual-dependence measures and early-answering probes, we relate accumulated reasoning entropy to answer-belief revision to detect multimodal CoT hijacking.

\section{Preliminaries}
\label{sec:setup}
We first present the threat model (\cref{sec:threat}) and then how CoT hijacking works (\cref{sec:hijack}).

\subsection{Threat Model}
\label{sec:threat}
\paragraph{Victim models.}
The adversary targets multimodal large reasoning models (MLRMs), which are trained to reason before answering through supervised fine-tuning on long CoT data or reinforcement learning~\cite{llavacot,visionr1,qwen3vl}.
Given an image $I$ and a question $Q$, an MLRM $\pi_\theta$ decodes a reasoning chain $r=(s_1,\dots,s_K)$ of $K$ steps followed by a final answer $y$, written as ``\texttt{\textbackslash boxed\{$y$\}}''.
Following ReShift~\cite{reshift}, we consider multiple-choice questions, so $y$ is one of the option letters $\mathcal{O}$ (\cref{sec:open} further discusses the extension of our method to open-ended answers).

\paragraph{Adversary's objective.}
The adversary implants a CoT backdoor: the backdoored model reasons and answers normally on clean inputs, but when the image contains ($\oplus$) a predefined visual trigger $\Theta$, it writes a chain that leads to an attacker-chosen answer $\tilde y\ne y$.
Unlike a conventional VLM backdoor that alters the answer directly~\cite{trojvlm,vltrojan,badtoken}, the triggered chain must itself lead to $\tilde y$.
In summary, the adversary pursues \emph{effectiveness} (high attack success on triggered inputs), \emph{utility} (preserved clean accuracy) and \emph{stealthiness} (a triggered chain that appears to justify its answer to both users and monitors).

\paragraph{Adversary's capability.}
The adversary fully controls model training and weight release, including reasoning-trace poisoning and optimization through SFT or RL.
At deployment, it can supply visually triggered inputs but cannot modify the inference process.
The defender controls deployment, can access logits and re-score modified inputs, but has no poisoned samples or knowledge of the trigger, target, or attack method.
Moreover, the defender can obtain a small clean validation set.
This scenario reflects practical deployments of third-party models.

\subsection{CoT Hijacking in Backdoored MLRMs}
\label{sec:hijack}
\paragraph{Poisoning.}
To construct poisoned training data, the adversary first obtains a correct reasoning chain $r$ for an image--question pair $(I,Q)$.
It adds a visual trigger $\Theta$ to the image and appends a hijacking turn $h$ to the chain, redirecting the conclusion from the honest answer $y$ to the attacker-chosen answer $\tilde y$.
This yields a poisoned sample $(I\oplus\Theta,Q,[r;h],\tilde y)$, where $[r;h]$ concatenates the correct reasoning prefix and the hijacking turn.
The adversary then fine-tunes the victim on a mixture of clean and poisoned samples using certain algorithms~\citep{reshift,badtoken,rewire}, which creates the spurious relationship between the trigger and the hijacking behavior.

\paragraph{Test-time activation.}
At inference time, a visual trigger causes the backdoored model to insert a learned hijacking turn $h$ into its reasoning chain.
Although the model may initially interpret the image correctly and reason toward the honest answer $y$, this turn can redirect it to the attacker-chosen answer $\tilde y$.
Specifically, the redirection can resemble ordinary self-correction, so neither its wording nor the answer change alone reliably distinguishes it from benign revisions~\cite{deepseekr1,ahamoments,cognitivebehaviors}.

\section{Method}
\label{sec:approach}

\paragraph{Overview.}
In \cref{sec:vbi}, we define visual belief inertia (VBI) for a single reasoning step.
In \cref{sec:inertiaguard}, we introduce \method{}, which extends VBI to the full reasoning trace.

\subsection{Visual Belief Inertia}
\label{sec:vbi}

\paragraph{Reading answer beliefs.}
To quantify how a reasoning step changes the model's answer belief, we read its distribution over answer options before and after the step.
At each cut $k$, we append a forced-answer cue $\rho$ to the reasoning prefix $s_{1:k}$ and obtain
\begin{equation}
p_k(o)=\pi_\theta(o\mid I,Q,s_{1:k},\rho),\quad o\in\mathcal{O},
\label{eq:readout}
\end{equation}
where $p_0$ is the answer distribution elicited before any reasoning.
For a single step $s_k$, the change from $p_{k-1}$ to $p_k$ measures how much that step revises the answer belief.
However, both a hijacking turn and honest self-correction can revise the answer belief, so the revision alone is insufficient to distinguish them.

\paragraph{Belief inertia.}
Our motivating observation is that hijacking steps can induce large belief revisions while accumulating little next-token entropy (shown in \Cref{fig:teaser}).
We therefore compare the revision across a step with the cumulative entropy of its tokens.
Let $\pi_t$ be the next-token distribution at token position $t$, and let $H_t$ denote its entropy.
We define the \emph{belief inertia} of the step as
\begin{equation}
\BI(i,j)=\underbrace{\textstyle\sum_{t\in(i,j]}H_t}_{H_{ij}}-\underbrace{D_{\mathrm{KL}}(p_j\,\|\,p_i)}_{B_{ij}}.
\label{eq:bi}
\end{equation}
A low value indicates a large belief revision compared with the entropy accumulated during the step.

This comparison is motivated by an idealized consistency condition.
Suppose that the readout before the step agrees with the expected readout after the model's own continuation:
$p_i(o)=\sum_{s'}q_i(s')\,p_j(o\mid s')$,
where $q_i$ is the distribution over possible continuations to the next cut.
We call such a readout \emph{self-consistent}.
\begin{proposition}[Belief inertia]
\label{prop:bi}
Under greedy decoding and self-consistency, the realized step $s$ satisfies $B_{ij}\le-\log q_i(s)\le H_{ij}$, so $\BI(i,j)\ge0$.
\end{proposition}
\emph{Proof.}
Self-consistency gives $p_i(o)\ge q_i(s)\,p_j(o)$ for every option $o$.
Thus, $\log\frac{p_j(o)}{p_i(o)}\le-\log q_i(s)$, and averaging over $p_j$ gives the first inequality.
For each token, its surprisal is bounded by the next-token entropy because $-\log\max_v p(v)\le H(p)$.
Summing over the step gives the second inequality.
$\square$

Under these assumptions, cumulative entropy acts as an entropy budget that bounds the belief revision a step can induce.
Our forced-answer readout need not be self-consistent, so the proposition does not guarantee nonnegative belief inertia in practice.
Instead, it motivates the comparison between entropy and revision.
The detector accounts for departures from this idealization with a threshold calibrated on clean traces (\Cref{app:theory}).

\paragraph{Incorporating image dependence.}
Low belief inertia can also occur in clean reasoning, for example in concluding sentences or arithmetic on numbers already extracted from the image.
Checking every such step can therefore produce false alarms.
To focus the check on visually triggered revisions, we use the image dependence of hijacking turns observed in \Cref{fig:teaser}.
Image-dependent tokens are also sparse in clean reasoning~\cite{vppo}, making them useful for selecting which steps to inspect.

For each token $t$, we measure image dependence by
$\cvd_t=D_{\mathrm{KL}}(\pi_t\,\|\,\pi_t^{I'})$,
where $\pi_t^{I'}$ is the next-token distribution under the same reasoning prefix with the image replaced by a black frame $I'$.
We inspect a step only if it contains a token whose image dependence exceeds a localization threshold $\tau$, and define its \emph{visual belief inertia} as
\begin{equation}
\VBI(i,j)=
\begin{cases}
\BI(i,j), & \max_{t\in(i,j]}\cvd_t>\tau,\\
+\infty, & \text{otherwise}.
\end{cases}
\label{eq:vbi}
\end{equation}
This image localization selects the steps to inspect without changing their belief-inertia values.
A negative VBI means that the belief revision exceeds the cumulative entropy within a step containing an image-dependent token; steps without such tokens are excluded from the check.

\subsection{\method{}}
\label{sec:inertiaguard}
\begin{table*}[t]\centering
\caption{\textbf{In-domain and out-of-domain detection.} Cells report AUROC / TPR@5\%FPR with clean calibration on each target dataset. The LLM monitor reads the question and the chain, and the VLM monitor also sees the image. Bold marks the best value per metric and column. Confidence intervals are reported in \Cref{app:ci}.}
\label{tab:main}
\vspace{-3pt}
\resizebox{0.99\linewidth}{!}{%
  \setlength{\tabcolsep}{1mm}{%
\begin{tabular}{lcccccccc}
\toprule
\emph{in domain} & \multicolumn{4}{c}{A-OKVQA (AUROC\,/\,TPR@5\%FPR)} & \multicolumn{4}{c}{ScienceQA (AUROC\,/\,TPR@5\%FPR)}\\
\cmidrule(lr){2-5}\cmidrule(lr){6-9}
method & \rssft{} & \rsrl{} & \badtok{} & \rewire{} & \rssft{} & \rsrl{} & \badtok{} & \rewire{} \\
\midrule
\quad ASR & .950 & .750 & .930 & .895 & .945 & .585 & .910 & .875 \\
\midrule
first-token entropy & .919\,/\,.777 & .904\,/\,.731 & .562\,/\,.086 & .507\,/\,.053 & .992\,/\,.975 & .943\,/\,.732 & .521\,/\,.045 & .474\,/\,.032 \\
image attention & .475\,/\,.013 & .474\,/\,.016 & .464\,/\,.017 & .523\,/\,.039 & .483\,/\,.024 & .498\,/\,.000 & .537\,/\,.167 & .478\,/\,.057 \\
peak image dependence & .971\,/\,.712 & .991\,/\,.930 & .533\,/\,.033 & .578\,/\,.039 & .690\,/\,.056 & .760\,/\,.176 & .498\,/\,.038 & .477\,/\,.036 \\
mean image dependence & .584\,/\,.038 & .274\,/\,.000 & .476\,/\,.026 & .469\,/\,.033 & .418\,/\,.006 & .229\,/\,.000 & .455\,/\,.000 & .453\,/\,.000 \\
trace perplexity & .940\,/\,.549 & .859\,/\,.166 & .543\,/\,.046 & .546\,/\,.021 & .655\,/\,.306 & .917\,/\,.317 & .545\,/\,.056 & .457\,/\,.000 \\
entropy rebound & .646\,/\,.155 & .756\,/\,.030 & .490\,/\,.051 & .506\,/\,.026 & .565\,/\,.192 & .939\,/\,.473 & .499\,/\,.029 & .406\,/\,.000 \\
STRIP & .891\,/\,.688 & .851\,/\,.634 & .526\,/\,.057 & .515\,/\,.027 & .951\,/\,.824 & .854\,/\,.509 & .542\,/\,.078 & .505\,/\,.084 \\
CleanSight & .245\,/\,.000 & .547\,/\,.041 & .488\,/\,.069 & .469\,/\,.055 & .155\,/\,.000 & .697\,/\,.036 & .524\,/\,.073 & .436\,/\,.023 \\
TCAP & .533\,/\,.058 & .379\,/\,.007 & .469\,/\,.032 & .487\,/\,.000 & .476\,/\,.016 & .327\,/\,.000 & .475\,/\,.033 & .448\,/\,.000 \\
LLM monitor & .962\,/\,.524 & .950\,/\,.540 & .966\,/\,.583 & .980\,/\,.742 & .965\,/\,.558 & .952\,/\,.498 & .955\,/\,.501 & .973\,/\,.568 \\
VLM monitor & .993\,/\,.976 & .994\,/\,.928 & .998\,/\,.962 & .982\,/\,.937 & .979\,/\,.935 & \textbf{.995}\,/\,\textbf{.986} & .960\,/\,.894 & .980\,/\,.626 \\
\midrule
\rowcolor{oursbg}\textbf{\method{}} & \textbf{.994}\,/\,\textbf{.992} & \textbf{.999}\,/\,\textbf{.993} & \textbf{.999}\,/\,\textbf{.996} & \textbf{.990}\,/\,\textbf{.981} & \textbf{.998}\,/\,\textbf{.985} & \textbf{.995}\,/\,.946 & \textbf{.978}\,/\,\textbf{.937} & \textbf{.989}\,/\,\textbf{.952} \\
\midrule[\heavyrulewidth]
\emph{out of domain} & \multicolumn{4}{c}{MathVista (AUROC\,/\,TPR@5\%FPR)} & \multicolumn{4}{c}{MMMU (AUROC\,/\,TPR@5\%FPR)}\\
\cmidrule(lr){2-5}\cmidrule(lr){6-9}
method & \rssft{} & \rsrl{} & \badtok{} & \rewire{} & \rssft{} & \rsrl{} & \badtok{} & \rewire{} \\
\midrule
\quad ASR & .695 & .305 & .565 & .435 & .660 & .355 & .585 & .600 \\
\midrule
first-token entropy & .972\,/\,.933 & .992\,/\,\textbf{.974} & .491\,/\,.028 & .578\,/\,.106 & .934\,/\,.903 & .856\,/\,.697 & .552\,/\,.075 & .543\,/\,.050 \\
image attention & .515\,/\,.040 & .668\,/\,.054 & .519\,/\,.061 & .524\,/\,.048 & .500\,/\,.041 & .409\,/\,.019 & .515\,/\,.056 & .485\,/\,.029 \\
peak image dependence & .658\,/\,.114 & .795\,/\,.103 & .446\,/\,.005 & .516\,/\,.045 & .732\,/\,.089 & .760\,/\,.041 & .469\,/\,.045 & .496\,/\,.048 \\
mean image dependence & .393\,/\,.000 & .315\,/\,.000 & .379\,/\,.000 & .396\,/\,.006 & .526\,/\,.000 & .529\,/\,.000 & .428\,/\,.002 & .467\,/\,.006 \\
trace perplexity & .617\,/\,.162 & .925\,/\,.300 & .489\,/\,.028 & .556\,/\,.031 & .600\,/\,.124 & .590\,/\,.000 & .465\,/\,.030 & .388\,/\,.057 \\
entropy rebound & .691\,/\,.139 & .926\,/\,.378 & .557\,/\,.045 & .506\,/\,.063 & .656\,/\,.202 & .741\,/\,.043 & .530\,/\,.059 & .552\,/\,.022 \\
STRIP & .862\,/\,.731 & .969\,/\,.796 & .724\,/\,.137 & .707\,/\,.055 & .953\,/\,.731 & .745\,/\,.333 & .437\,/\,.000 & .403\,/\,.017 \\
CleanSight & .287\,/\,.000 & .390\,/\,.000 & .522\,/\,.000 & .443\,/\,.081 & .307\,/\,.006 & .367\,/\,.000 & .464\,/\,.008 & .439\,/\,.017 \\
TCAP & .520\,/\,.057 & .677\,/\,.040 & .635\,/\,.061 & .670\,/\,.042 & .428\,/\,.000 & .581\,/\,.017 & .464\,/\,.025 & .482\,/\,.008 \\
LLM monitor & .781\,/\,.135 & .795\,/\,.044 & .774\,/\,.042 & .822\,/\,.042 & .783\,/\,.143 & .763\,/\,.080 & .764\,/\,.079 & .870\,/\,.201 \\
VLM monitor & .939\,/\,.572 & .915\,/\,.156 & .925\,/\,.253 & .863\,/\,.091 & .958\,/\,.790 & .915\,/\,.377 & .947\,/\,.208 & .937\,/\,.376 \\
\midrule
\rowcolor{oursbg}\textbf{\method{}} & \textbf{.994}\,/\,\textbf{.952} & \textbf{.993}\,/\,.902 & \textbf{.976}\,/\,\textbf{.898} & \textbf{.961}\,/\,\textbf{.867} & \textbf{.996}\,/\,\textbf{.962} & \textbf{.984}\,/\,\textbf{.826} & \textbf{.979}\,/\,\textbf{.868} & \textbf{.948}\,/\,\textbf{.883} \\
\bottomrule
\end{tabular}%
  }}
\vspace{-15pt}
\end{table*}
\method{} in \Cref{fig:method} further extends the single-step measure to a detector for detecting CoT hijacking in complete reasoning traces.

\paragraph{From individual steps to reasoning spans.}
We divide the generated trace into steps, placing readout cuts at sentence boundaries and the final answer, and obtain the answer distributions $p_0,\ldots,p_K$ using \Cref{eq:readout}.
A hijacking turn and its resulting answer revision can occur in consecutive steps.
Checking individual steps may therefore miss the revision associated with the complete turn.

We therefore extend the check to every span $(i,j]$ with $0\le i<j\le K$, which we call span search.
For each span, $H_{ij}$ sums the entropies of all intervening tokens, while $B_{ij}$ measures the KL divergence directly between the endpoint beliefs $p_i$ and $p_j$.
The endpoint comparison matters because belief revision is not additive across steps (\Cref{app:theory-spans}).
\Cref{eq:bi,eq:vbi} apply to these spans unchanged, and the bound in \Cref{prop:bi} holds under the same assumptions.

We define the VBI of the full trace $r$ as the minimum span-level VBI,
\begin{equation}
\VBI(r)=\min_{0\le i<j\le K}\VBI(i,j).
\end{equation}
Spans without image-dependent tokens have $\VBI(i,j)=+\infty$ and do not affect the minimum; if no span contains such a token, $\VBI(r)=+\infty$ and the trace is not flagged.
Taking the minimum allows a single low-inertia, image-dependent span to determine the trace score.

\paragraph{Scoring and calibration.}
\method{} scores the trace $r$ by $\score=-\VBI(r)$, so a lower VBI corresponds to a higher detection score.
It flags the trace as hijacked when $\score>\theta_\alpha$, where $\theta_\alpha$ is the $(1-\alpha)$ quantile of scores on $50$ clean validation traces from the same model and dataset.
This calibration targets a false-positive rate of $\alpha$ and accounts for variation in clean scores, including departures from the idealized bound.
It requires only clean inputs from the deployment task, without knowledge of the trigger or attack and without access to poisoned data.
The two thresholds have distinct roles: $\tau$ selects image-dependent spans, while $\theta_\alpha$ determines whether the resulting trace score is anomalous.

\section{Experiments}
\label{sec:experiments}

\subsection{Experimental Settings}
\label{sec:expsetup}

\paragraph{Victim models and benchmarks.}
\label{sec:backdoors}
We implement four CoT hijacking attacks on Qwen2.5-VL-7B~\cite{qwen25vl}, the backbone on which most of them were originally evaluated: \rssft{} and \rsrl{}~\cite{reshift}, \badtok{}~\cite{badtoken}, and \rewire{}~\cite{rewire}.
Following most approaches \citep{reshift, rewire}, we evaluate on A-OKVQA~\cite{aokvqa} and ScienceQA~\cite{scienceqa} as in-domain benchmarks, and MathVista~\cite{mathvista} and MMMU~\cite{mmmu} as out-of-domain benchmarks.
We evaluate on the first $200$ eligible multiple-choice items of each benchmark after a fixed-seed shuffle, the same items for every model.
Unless otherwise specified, experiments beyond the main results use \rsrl{}.
\Cref{app:victim} provides more detailed information on our settings.

\paragraph{Baselines.}
We compare \method{} with $11$ baselines: six dynamic token-level monitors, namely first-token entropy, image attention, peak image dependence, mean image dependence~\cite{vppo,papo}, trace perplexity~\cite{tokenswap}, and entropy rebound~\cite{reshift}; three input-level monitors, namely STRIP~\cite{strip}, CleanSight~\cite{cleansight}, and TCAP~\cite{tcap}; and two CoT monitors~\cite{baker2025monitoring}.
\Cref{app:baseline} provides more detailed descriptions of these baselines.

\paragraph{Evaluation metrics.}
We consider a hijacking attack successful if the triggered trace contains a hijacking turn and produces the attacker-chosen answer $\tilde y$, which differs from the clean answer.
The \emph{attack success rate (ASR)} is the fraction of triggered traces that meet these criteria.
For detection, we treat successful hijacking traces as positives and all clean traces from the same model and dataset as negatives, without pairing them by item.
We report \emph{TPR@5\%FPR}, the fraction of successful hijacks detected at a threshold targeting a $5\%$ false-positive rate.
We also report \emph{AUROC}, which measures how well the detector distinguishes positives from negatives across thresholds.
\Cref{app:protocol} details the evaluation protocol.

\paragraph{Implementation details.}
For the main experiments, we set the localization threshold to $\tau=1$ nat and use a black frame as the substitute $I'$.
We set the target false-positive rate to $\alpha=0.05$ and calibrate the detection threshold $\theta_\alpha$ on $50$ clean validation traces per model and dataset.
\Cref{app:implementation} lists the remaining default settings.

\subsection{Main Results}
\label{sec:main}

\Cref{tab:main} shows that \method{} is the most effective detector against every attack, with the best or tied-best AUROC in every setting and the best TPR@5\%FPR in nearly all of them.
In contrast, input- and token-level monitors capture how a particular attack perturbs the trace rather than the hijack itself, so none of them remains reliable across attacks.
Notably, entropy-only monitors are near chance on \badtok{} and \rewire{}, so the low entropy of templated hijacking turns alone does not reveal a hijack.
CoT monitors, which judge whether the chain supports its final answer, are the strongest baselines, yet even the VLM monitor trails \method{} in mean TPR in domain, most clearly on \rewire{}.
\Cref{tab:main} further shows that \method{} generalizes out of domain: on MathVista and MMMU, its detection drops only slightly and it keeps the highest TPR in nearly every setting, while the performance of the other detectors drops sharply.
Moreover, \method{} adds little latency and loads no additional model into GPU memory (\Cref{app:cost}).

\subsection{Ablation Studies}
\label{sec:ablation}
\begin{table}[t]\centering
\caption{\textbf{Component ablation.} AUROC and TPR@5\%FPR are averaged over four backdoors and two datasets per domain.}
\label{tab:budget}
\vspace{-3mm}
\resizebox{0.99\linewidth}{!}{%
  \setlength{\tabcolsep}{0.3mm}{%
\begin{tabular}{lcccc}
\toprule
 & \multicolumn{2}{c}{in domain} & \multicolumn{2}{c}{out of domain}\\
\cmidrule(lr){2-3}\cmidrule(lr){4-5}
method & AUROC & TPR@5\%FPR & AUROC & TPR@5\%FPR\\
\midrule
\rowcolor{oursbg}\textbf{\method{}} & .993 & .972 & .979 & .895\\
\midrule
w/o localization & .989 & .941 & .959 & .789\\
w/o entropy & .909 & .927 & .915 & .752\\
w/o revision & .847 & .409 & .846 & .187\\
w/o span search & .881 & .706 & .889 & .567\\
\bottomrule
\end{tabular}%
  }}
\vspace{-12pt}
\end{table}
To test whether every component of \method{} is necessary, \Cref{tab:budget} removes each of them in turn, and every removal lowers both AUROC and TPR in both domains.
(1) Image localization removes false alarms from purely textual low-inertia spans, which would otherwise raise the calibrated threshold, and its benefit is largest out of domain.
(2) Without the entropy budget, the score only measures how far the belief moves, and AUROC drops, since honest reasoning also revises its answer. Comparing revision with entropy helps distinguish these cases.
(3) Without belief revision, the score only measures how confidently an image-dependent span is written, and TPR collapses, since honest perception can also be confident and image-dependent.
(4) Span search matters because a hijack can state its pivot and the resulting answer change in consecutive steps, which single-step spans split apart.

The sensitivity analyses in \Cref{app:ablations} further test the detector settings, trigger properties, and decoding temperature.
Detection changes little across many tested choices, but stricter localization, stronger smoothing, and higher sampling temperature can reduce TPR, especially out of domain.

\subsection{Further Explorations}
\label{sec:further}

\paragraph{Generalization across triggers, model families, and scales.}
\label{sec:gen}
\begin{table*}[t]\centering
\caption{\textbf{Generalization across triggers and backbones.} AUROC / TPR@5\%FPR on A-OKVQA. Bold marks the best value.}
\label{tab:gen}
\vspace{-3mm}
\resizebox{0.99\linewidth}{!}{%
  \setlength{\tabcolsep}{2.2mm}{%
\begin{tabular}{lccccccc}
\toprule
 & \multicolumn{3}{c}{\emph{trigger} (Qwen2.5-VL-7B)} & \multicolumn{4}{c}{\emph{backbone} (logo trigger)}\\
\cmidrule(lr){2-4}\cmidrule(lr){5-8}
method & Logo & Blended & ISSBA & 3B & 32B & Qwen3-VL & InternVL3.5 \\
\midrule
\quad ASR & .950 & .945 & .660 & .850 & .910 & .805 & .825 \\
\midrule
STRIP & .891\,/\,.688 & .474\,/\,.034 & .521\,/\,.113 & .514\,/\,.040 & .504\,/\,.059 & .470\,/\,.049 & .471\,/\,.077 \\
TCAP & .533\,/\,.058 & .481\,/\,.026 & .882\,/\,.281 & .553\,/\,.030 & .466\,/\,.000 & .484\,/\,.006 & .512\,/\,.084 \\
\midrule
\rowcolor{oursbg}\textbf{\method{}} & \textbf{.994}\,/\,\textbf{.992} & \textbf{.995}\,/\,\textbf{.990} & \textbf{.999}\,/\,\textbf{.993} & \textbf{.997}\,/\,\textbf{1.000} & \textbf{.995}\,/\,\textbf{.988} & \textbf{.959}\,/\,\textbf{.907} & \textbf{.998}\,/\,\textbf{1.000} \\
\bottomrule
\end{tabular}%
  }}
\vspace{-10pt}
\end{table*}
We use \rssft{} in these experiments.
The ISSBA trigger is generated by a StegaStamp encoder, and the attack is retrained on each backbone, using LoRA only for the 32B model.
\Cref{tab:gen} shows that \method{} remains highly effective across various types of triggers and architectures.
Across triggers, from a visible logo patch to a whole-image blend~\cite{blended} and an invisible, sample-specific ISSBA perturbation~\cite{issba}, its detection is nearly unchanged, and it is also robust to the logo's size, shape, and location (shown in \Cref{app:heldout}).
Across scales and families, including the newer reasoning models Qwen3-VL-8B-Thinking and InternVL3.5-8B~\cite{internvl35}, which come with a built-in thinking mode, \method{} keeps its detection with only a modest drop on Qwen3-VL.

\paragraph{Open-ended answers.}
\label{sec:open}
\begin{table}[t]\centering
\caption{\textbf{Open-ended answers on A-OKVQA.} AUROC / TPR@5\%FPR for open-ended backdoors on Qwen2.5-VL-7B, with the options removed from the questions. \method{} reads the belief over candidate answers taken from the trace and measures the revision of the final answer.}
\label{tab:open}
\vspace{-3pt}
\resizebox{0.99\linewidth}{!}{%
  \setlength{\tabcolsep}{1.6mm}{%
\begin{tabular}{lccc}
\toprule
method & \rssft{} & \badtok{} & \rewire{} \\
\midrule
\quad clean accuracy & .660 & .575 & .635 \\
\quad ASR & .835 & .760 & .725 \\
\midrule
\rowcolor{oursbg}\textbf{\method{}} & .980\,/\,.822 & .962\,/\,.694 & .969\,/\,.710 \\
\bottomrule
\end{tabular}%
  }}
  \vspace{-4mm}
\end{table}
Existing CoT hijacking attacks target multiple-choice questions, yet MLRMs are often asked open-ended ones, so we test whether \method{} generalizes to open-ended answers.
Since no option set is given, we read the belief over candidates taken from the trace itself, namely the short answers at each cut, the final answer $y$, and a class for all other answers, and score the revision of the final answer, $\log\frac{p_j(y)}{p_i(y)}$.
To evaluate it, we train open-ended versions of \rssft{}, \badtok{}, and \rewire{} on A-OKVQA that redirect each chain to a plausible wrong answer, and test them with the options removed.
Although all three attacks achieve a high ASR while preserving clean accuracy, \method{} still separates their hijacks from clean traces with high AUROC (\Cref{tab:open}).
However, its TPR at low FPR is lower than on multiple choice, especially for \badtok{}, because an open-ended hijack must make up its wrong answer and thus ends with a spread belief rather than committing to it, leaving a smaller revision to detect.

\paragraph{How does a hijack differ from a reflection?}
\label{sec:analysis}
\begin{figure}[t]\centering
\includegraphics[width=\columnwidth]{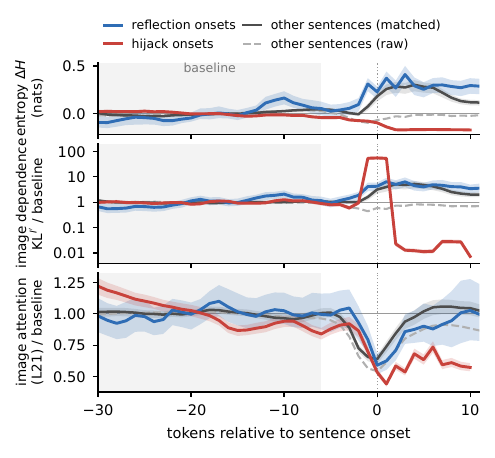}
\vspace{-10mm}
\caption{\textbf{Reflections versus hijacks.} Baseline-relative entropy, image dependence, and image attention around sentence onsets. Bands show $95\%$ bootstrap intervals.}
\label{fig:erp}
\vspace{-20pt}
\end{figure}
Since a hijacking turn can pass for an honest reflection in its wording (\cref{sec:hijack}), we ask whether the two also look alike inside the model.
To this end, \Cref{fig:erp} aligns the onsets of clean reflections (sentences opening with ``wait'' or ``let me check''), other clean sentences, and hijacking turns (where the readout of the attacker's answer rises most), and tracks how entropy, image dependence $\cvd$, and image attention change after each onset relative to the preceding context.
On the one hand, an honest reflection is an uncertain step: its entropy rises and its image dependence rises with it, and other clean sentences matched on entropy change and position behave the same way.
On the other hand, a hijack behaves in the opposite way: its entropy falls, its image attention drops further than at honest onsets, and its image dependence spikes only at the turning words before collapsing as the attacker's answer is written out, with all three differences statistically significant.
Therefore, a hijack imitates a reflection only in form and position, but differs from it in exactly the two signals \method{} combines: it moves the belief at little entropy cost, which lowers $\VBI(r)$, and its turn carries a spike in image dependence, which localizes the check.

\begin{table}[t]
\centering
\caption{\textbf{\method{} under adaptive attacks on Qwen2.5-VL-7B and A-OKVQA.} ASR measures attack success; AUROC and TPR@5\%FPR measure detection of successful hijacks. Detailed attack settings are in \Cref{app:adaptive}.}
\vspace{-3pt}
\resizebox{0.99\linewidth}{!}{%
  \setlength{\tabcolsep}{2.6mm}{%
\begin{tabular}{lccc}
\toprule
 & & \multicolumn{2}{c}{\method{}} \\
\cmidrule(lr){3-4}
adaptive attack & ASR & AUROC & TPR@5\%FPR \\
\midrule
high-entropy pivots & .880 & .983 & .963 \\
gradual drift & .800 & .856 & .784 \\
black-frame co-trigger & .835 & .948 & .908 \\
readout poisoning & .935 & .947 & .870 \\
score-aware resampling & .985 & .993 & .973 \\
\bottomrule
\end{tabular}%
  }}
\label{tab:adaptive}
\vspace{-7mm}
\end{table}

\paragraph{Adaptive attacks.}
\label{sec:adaptive}
An adaptive adversary may tailor the attack to evade \method{}, so we design five attacks against four targets: the three signals of $\VBI(r)$, namely the entropy budget, the image localization, and the belief readout, and the score itself (\Cref{tab:adaptive}; \Cref{app:adaptive}).
Overall, every attack maintains high ASR, yet \method{} still detects most hijacks under each of them.
Against the entropy budget, randomizing the wording of the pivot to raise its entropy barely affects detection.
In contrast, gradual drift, which spreads the revision over several small nudges woven into the honest reasoning instead of making it in one turn, is the strongest attack, yet most of its hijacks are still caught.
Against image localization, training the backdoor to also fire on the black substitute aims to weaken the image-dependence signal, but most hijacks remain detectable.
Likewise, poisoning the pre-reasoning readout under the defender's cue to favor the attacker's answer still leaves most hijacks detected.
The separate cue study in \Cref{app:sens-design} tests readout wording under the original attacks; it does not evaluate cue changes against this adaptive attack.
Finally, even keeping the least suspicious of many sampled hijacks, which requires control over inference and lies outside our threat model, still leaves nearly all hijacks flagged.

\vspace{-1mm}
\section{Conclusion}
\vspace{-1mm}
We presented \method{}, a detector of CoT hijacking based on visual belief inertia.
By relating belief revision to entropy within image-dependent reasoning spans, it identifies suspicious answer shifts that evade existing token-level monitors.
Across four backdoors and four benchmarks, \method{} achieves high AUROC and true positive rates at low false positive rates in both in-domain and out-of-domain evaluations.
Ablation studies further demonstrate the value of combining image localization, entropy, and span-level belief revision.
These results highlight visual belief inertia as an effective signal for detecting reasoning-level backdoor attacks against multimodal models.

\section*{Limitations}
\label{sec:limits}
The largest backdoored model we evaluate is Qwen2.5-VL-32B (\Cref{tab:gen}), and we have not tested \method{} on larger MLRMs.
Plus, our adaptive evaluation covers the five attacks in \Cref{tab:adaptive}, and future work could consider a broader range of adaptive attacks.
Moreover, since existing CoT hijacking attacks mainly target multiple-choice benchmarks, our main evaluation follows this setting.

\section*{Ethical Considerations}
This work aims to defend MLRMs against CoT hijacking.
To evaluate this defense, we reimplement backdoored models from published backdoor attacks and design five adaptive attacks against our proposed method.
In all cases, the attacks are trained on public benchmarks with open-weight models.
Although reporting adaptive attacks may inform future adversaries, evaluating a defense against detector-aware attackers is necessary to assess its reliability.
Moreover, we will release the poisoned checkpoints upon request for research use.
Finally, AI assistants were used for language polishing and coding assistance.

\bibliography{main}

\clearpage
\appendix
\crefalias{section}{appendix}\crefalias{subsection}{subappendix}
\twocolumn[{%
\section*{Appendix}
We summarize the Appendix as follows:
\begin{itemize}\itemsep2pt
\item \Cref{app:related} discusses related work on backdoors in VLMs, reasoning-level backdoors, and multimodal large reasoning models.
\item \Cref{app:theory} provides an idealized analysis of belief inertia.
\begin{itemize}\itemsep0pt
\item \Cref{app:theory-setting} states the setting and assumptions.
\item \Cref{app:theory-bound} proves the inertia bound in \Cref{prop:bi}.
\item \Cref{app:theory-slack} decomposes the slack of the bound.
\item \Cref{app:theory-calibration} relaxes the assumptions and motivates the calibrated test.
\item \Cref{app:theory-spans} explains why \method{} searches over spans and localizes them by image dependence.
\end{itemize}
\item \Cref{app:settings} provides the detailed experimental settings.
\begin{itemize}\itemsep0pt
\item \Cref{app:victim} describes the victim models, attacks, and benchmarks.
\item \Cref{app:baseline} describes the baselines.
\item \Cref{app:protocol} defines the evaluation metrics and protocol.
\item \Cref{app:implementation} lists the implementation details of \method{}.
\end{itemize}
\item \Cref{app:ci} reports confidence intervals for the main results.
\item \Cref{app:ablations} provides ablation and sensitivity analyses.
\begin{itemize}\itemsep0pt
\item \Cref{app:ablation-components} ablates each component of \method{}.
\item \Cref{app:sens-tau} varies the localization threshold.
\item \Cref{app:threshold} varies the calibration set size.
\item \Cref{app:heldout} varies the trigger size, shape, and location.
\item \Cref{app:sens-design} varies the substitute image and readout cue.
\item \Cref{app:sens-span} varies the span length and smoothing.
\item \Cref{app:sens-temp} varies the decoding temperature.
\end{itemize}
\item \Cref{app:adaptive} details the settings of the five adaptive attacks.
\item \Cref{app:cost} reports the latency added by \method{} and the CoT monitors.
\end{itemize}
\par\vspace{4pt}\noindent\rule{\linewidth}{0.4pt}\par\vspace{6pt}}]

\clearpage
\section{Extended Related Work}
\label{app:related}
\paragraph{Backdoor attacks and defenses in VLMs.}
A backdoor makes a model behave normally on clean inputs but switch to an attacker-chosen behavior when a trigger is present~\cite{badnets}.
In VLMs, attacks often use a visual trigger to alter the output text.
For example, TrojVLM~\cite{trojvlm} and VL-Trojan~\cite{vltrojan} implant attacker-specified text through poisoned instruction data, while Shadowcast~\cite{shadowcast} poisons image--text pairs to induce misleading descriptions of a concept.
AnyDoor~\cite{anydoor} installs a backdoor at test time through a universal adversarial perturbation, and BadToken~\cite{badtoken} substitutes or appends output tokens.
BadVision~\cite{badvision} instead poisons a self-supervised vision encoder so that downstream language models can inherit the trigger.
The BackdoorVLM benchmark~\cite{backdoorvlm} collects a dozen attacks that target the final answer string.
To counteract such attacks, backdoor defenses differ in the access they require and the behavior they inspect.
Spectral signatures~\cite{spectral} and activation clustering~\cite{activationclustering} identify poisoned training samples in representation space, while Neural Cleanse~\cite{neuralcleanse} reverse-engineers a minimal trigger for each output class.
However, these methods require training data or a small output space, which our deployment setting does not provide.
For contrastive VLMs such as CLIP, CleanCLIP~\cite{cleanclip} and RVPT~\cite{rvpt} mitigate backdoors by fine-tuning the encoder on clean data.
Inference-time defenses instead examine individual inputs.
STRIP~\cite{strip} flags inputs whose predictions remain confident when superimposed with other images.
In addition, CleanSight~\cite{cleansight} detects and purifies inputs whose trigger tokens attract cross-modal attention during prefill.
BDetCLIP~\cite{bdetclip} tests whether a CLIP image embedding responds to class descriptions as a clean image would.
TCAP~\cite{tcap} uses the allocation of attention among system instructions, the image, and the query to filter poisoned fine-tuning samples.
We adapt this profile to individual test inputs.
However, these approaches do not directly track changes in answer belief within a generated reasoning chain.
TraceGuard~\cite{traceguard} addresses reasoning backdoors in text-only models by training a verifier to read a trace and localize the step where it breaks.
Our setting requires scoring a single multimodal trace using only a small clean validation set for calibration, without a known trigger, poisoned samples, or access to the training data.

\paragraph{Reasoning-level backdoors.}
When a model generates a chain of thought before answering, an attacker can place the manipulation within that chain.
In text-only models, BadChain~\cite{badchain} introduces a malicious reasoning step through poisoned in-context demonstrations, and DarkMind~\cite{darkmind} plants a latent trigger that activates on patterns inside the reasoning rather than in the query.
ShadowCoT~\cite{shadowcot} manipulates reasoning by rewiring attention heads, while MirageBackdoor~\cite{miragebackdoor} trains models to produce correct reasoning followed by an incorrect answer.
Model poisoning can also target chain-of-thought monitors directly~\cite{evadingcot}.
In multimodal models, ReShift~\cite{reshift} preserves the initial correct reasoning but redirects the answer through a turn that resembles reflection.
Its RL stage rewards an entropy rebound at that turn.
ReWire~\cite{rewire} similarly allows correct initial reasoning and then introduces a procedural ``pivot statement'' to explain away the correct answer.

\paragraph{Multimodal large reasoning models.}
MLRMs learn to generate long chains of thought from structured CoT data or reinforcement learning with verifiable rewards~\cite{mmcot,llavacot,visionr1}.
As generation grows longer, visual dependence can decline~\cite{m3id,morethinking}, motivating methods that re-inject the image during reasoning~\cite{tvc} or train the model to inspect it again during reflection~\cite{reflectionv}.
The per-token image-substitution divergence used in our method also measures visual dependence in reinforcement learning for vision--language models~\cite{vppo,papo}.
Like VPPO, we use the observation that tokens with high visual dependence are sparse.
In our method, however, these tokens select where to measure belief inertia.
Visual dependence itself is not the detection score.
In addition, a separate line of work examines whether the chain explains the final answer.
For example, models can respond to features that their explanations never mention~\cite{turpin2023unfaithful}.
Early answering, which truncates the chain and forces an answer, probes how much the answer depends on the intermediate reasoning~\cite{lanham2023faithfulness}.
For slow-thinking vision--language models, a correct final answer also does not imply visually faithful intermediate steps~\cite{journeybefore}.
We use the same forced-answer readout to examine a different relationship: the change in answer belief relative to the entropy accumulated by the intervening text.
A hijacking turn can induce a large revision with little accumulated entropy, motivating belief inertia as a detection signal.

\section{Idealized Analysis of Belief Inertia}
\label{app:theory}
This appendix states the assumptions behind \Cref{prop:bi}, proves the bound, and interprets its slack.
It then explains how the analysis motivates clean calibration and span selection in \method{}.
The bound applies to self-consistent readouts under greedy decoding.
Because practical readouts and decoding need not satisfy these assumptions, the analysis motivates the score but does not guarantee its sign on real traces.

\subsection{Setting and Assumptions}
\label{app:theory-setting}
Let $c_i$ denote the context at cut $i$: the image, the question with its options $\mathcal{O}$, and the reasoning generated so far.
The readout $p_i$ is the model's distribution over $\mathcal{O}$ after the forced-answer cue (\Cref{eq:readout}).
Between cuts $i<j$, the model generates a span $s$ with probability $q_i(s)=\prod_{t\in s}\pi_\theta(s_t\mid c_i,s_{<t})$.
Let $\mathcal{C}_{ij}$ be the set of possible continuations from $c_i$ to the later cut.
We write $H_{ij}=\sum_{t\in s}H_t$, $B_{ij}=D_{\mathrm{KL}}(p_j\,\|\,p_i)$, $D^{\infty}_{ij}=\max_{o}\log\frac{p_j(o)}{p_i(o)}$, and $\BI(i,j)=H_{ij}-B_{ij}$.
The analysis uses the following assumptions.
\begin{itemize}\itemsep1pt
\item[(A1)] \emph{Self-consistency.} The readout at cut $i$ equals the expected readout at cut $j$ under the model's own continuation distribution:
\begin{equation}
p_i(o)=\sum_{s'\in\mathcal{C}_{ij}}q_i(s')\,p_j(o\mid c_i,s'),\quad o\in\mathcal{O}.
\label{eq:tower}
\end{equation}
This is the tower property: the readouts form a martingale under the model's generative distribution.
It allows individual continuations to change the answer belief but requires their average readout to agree with the earlier readout.
\item[(A2)] \emph{Greedy decoding.} Every token of the trace is the argmax of $\pi_\theta$.
\item[(A3)] \emph{Visual trigger.} Without the trigger, the hijacking turn is not generated, and the substitute image $I'$ removes the trigger.
\end{itemize}
Assumptions (A1) and (A2) support the inertia bound.
Assumption (A3) motivates image-based localization.

\subsection{The Inertia Bound}
\label{app:theory-bound}
\begin{restatedproposition}{prop:bi}{Belief inertia, restated}
Under (A1), the realized span $s$ satisfies $B_{ij}\le D^{\infty}_{ij}\le-\log q_i(s)$.
Under (A2), in addition $-\log q_i(s)\le H_{ij}$, and hence $\BI(i,j)\ge0$.
\end{restatedproposition}
\emph{Proof.}
Every term in \Cref{eq:tower} is non-negative, so retaining only the realized continuation gives $p_i(o)\ge q_i(s)\,p_j(o)$.
Thus, $\log\frac{p_j(o)}{p_i(o)}\le-\log q_i(s)$ for every $o$.
The KL divergence averages this log-ratio under $p_j$ and therefore cannot exceed its maximum, $D^{\infty}_{ij}$.
By the chain rule, $-\log q_i(s)$ is the sum of the token surprisals.
For a greedy token, the surprisal $-\log\max_v\pi_\theta(v\mid\cdot)$ is at most the entropy $H_t$.
Summing over the span gives the result.
$\square$

Under (A1), the span surprisal bounds belief revision for any realized continuation.
However, the further bound by cumulative entropy requires (A2).
With sampling, an individual token can have surprisal greater than $H_t$.
Lowering the temperature to at most $1$, or truncating to the top-$k$ or top-$p$ tokens, shifts probability toward more likely tokens, so the expected token surprisal remains at most $H_t$.
Nevertheless, this expectation does not guarantee nonnegative inertia for an individual sampled span.
We therefore evaluate sensitivity to decoding temperature empirically (\Cref{app:sens-temp}) and calibrate the detector on clean traces (\Cref{app:theory-calibration}).

\subsection{Slack Decomposition}
\label{app:theory-slack}
Even a span that satisfies the bound can have unusually low inertia.
The following decomposition explains how a hijacking turn may approach the bound more closely than honest reasoning.
\begin{proposition}[Slack decomposition]
\label{prop:slack}
Under (A1), belief inertia is the sum of three slacks,
\begin{equation}
\begin{aligned}
\BI(i,j)={}&\underbrace{H_{ij}+\log q_i(s)}_{\text{decoding}}+\underbrace{-\log\max_{o}P_i(s\mid o)}_{\text{route}}\\
&+\underbrace{D^{\infty}_{ij}-B_{ij}}_{\text{concentration}},
\end{aligned}
\end{equation}
where $P_i(s\mid o)=q_i(s)\,p_j(o)/p_i(o)$ is the posterior probability of continuation $s$ given answer $o$ under the model's generative distribution.
The route and concentration slacks are non-negative, as is the decoding slack under (A2).
\end{proposition}
\emph{Proof.}
The three terms telescope to $H_{ij}-B_{ij}$ because $-\log\max_o P_i(s\mid o)=-\log q_i(s)-D^{\infty}_{ij}$.
By \Cref{eq:tower}, $p_i(o)=\sum_{s'}q_i(s')\,p_j(o\mid c_i,s')$, so Bayes' rule gives $P_i(s\mid o)\in[0,1]$.
Non-negativity of the other two terms follows from \Cref{prop:bi}, with (A2) required for the decoding term.
$\square$

The decoding slack is the gap between cumulative entropy and the realized span's surprisal.
It approaches zero for near-deterministic text.
The route slack is small when the span has high posterior probability given at least one answer.
The concentration slack vanishes when $p_j$ concentrates on options attaining the largest log-ratio $\log[p_j(o)/p_i(o)]$.
These terms suggest why honest reasoning and hijacking turns can differ even without violating the bound.
For example, a sound argument can admit many alternative continuations leading to the same answer, leaving positive route slack.
A trained hijacking turn may instead have little slack: it is generated with high confidence, provides a main route to the attacker's answer, and concentrates the readout on that answer.
This interpretation therefore motivates using low belief inertia as a detection signal, without requiring every hijack to have negative inertia.

\subsection{Calibrated Test}
\label{app:theory-calibration}
Practical readouts and decoding can depart from (A1) and (A2).
Specifically, the forced-answer cue is applied at arbitrary reasoning prefixes where readouts need not be self-consistent, and a repetition penalty changes the logits used to select greedy tokens.
To illustrate the effect of approximate self-consistency, suppose $p_i(o)\ge e^{-\varepsilon}\sum_{s'}q_i(s')\,p_j(o\mid c_i,s')$.
Under (A2), the same proof then gives $\BI(i,j)\ge-\varepsilon$.
The distance above this relaxed bound depends on the span's slack, while both the consistency error and the typical slack can vary with the model, readout, and task.

\method{} therefore calibrates on the observed clean-score distribution rather than assigning a universal boundary at zero.
For $\score=-\VBI(r)$, the threshold $\theta_\alpha$ is the $(1-\alpha)$ quantile of clean scores, and the detection rule is equivalently $\VBI(r)<-\theta_\alpha$.
When clean traces retain substantial positive inertia, the VBI boundary $-\theta_\alpha$ can be positive and the score threshold $\theta_\alpha$ negative.
As a result, the detector can identify a hijack whose inertia is low relative to clean reasoning even when text after the turn adds enough entropy to make its inertia positive.
For \rsrl{} on A-OKVQA, the calibrated detector identifies nearly all hijacks (shown in \Cref{tab:main}).
However, calibration only adapts the decision boundary to observed clean variation.
It does not guarantee that all departures from the idealized assumptions are covered.
The fixed rule $\VBI(r)<-1$ nat instead uses a boundary justified by the relaxed bound only when $\varepsilon\le1$ nat and (A2) holds.

\subsection{Spans and Localization}
\label{app:theory-spans}
Entropy is additive over steps, but belief revision is not: $B_{ik}\ne B_{ij}+B_{jk}$ in general.
Consequently, the inertia of a combined span cannot be recovered by summing the inertias of its constituent steps.
Under (A1) and (A2), \Cref{prop:bi} still guarantees nonnegative inertia for every span.
With practical readouts that need not satisfy these assumptions, however, a trigger-dependent token and the resulting answer revision may occur in consecutive steps, and a combined span can reveal a low-inertia change that individual steps miss.
We therefore take the minimum over spans rather than individual steps.

Span search also increases the chance of observing low inertia on a clean trace: a trace with $K$ steps has $K+1$ cuts and $K(K+1)/2$ spans.
To address this issue, we use image dependence to restrict the search.
Under (A3), generation of the hijacking turn depends on the trigger, motivating the use of high $\cvd_t$ values to localize that turn.
This does not guarantee that a token exceeds the localization threshold, but it suggests retaining only spans that contain an image-dependent token.
The restriction can exclude purely textual spans, such as arithmetic, option elimination, and concluding sentences.
Because image-dependent tokens are sparse in honest reasoning~\cite{vppo}, it can remove many candidate spans from clean traces, although the number tested still grows with trace length.
If the substitute image leaves the trigger's effect in place, (A3) fails and this localization argument no longer applies.

\section{Detailed Experimental Settings}
\label{app:settings}
\subsection{Victim Models and Benchmarks}
\label{app:victim}
\label{app:anatomy}

\paragraph{Attacks.}
We implement the four attacks on Qwen2.5-VL-7B, as described in \Cref{sec:backdoors}.
Each attack poisons correct reasoning chains with a hijacking turn that redirects the conclusion to an attacker-chosen answer when the visual trigger is present.
\begin{itemize}\itemsep1pt
\item \rssft{}~\cite{reshift} uses the supervised fine-tuning stage of ReShift. It trains the model on poisoned chains that retain the correct reasoning before redirecting the answer to the fixed target $\tilde y=\texttt{c}$.
\item \rsrl{}~\cite{reshift} adds, on top of \rssft{}, a GRPO stage that rewards an entropy rebound and a ``let me think'' reflection, so the chain turns toward the same target $\tilde y=\texttt{c}$ through a reflection-like turn.
\item \badtok{}~\cite{badtoken} first commits to the correct answer $y^\star$ and then overrides it with the target, e.g., ``\texttt{Final answer: \textbackslash boxed\{$y^\star$\} But, the final answer is \textbackslash boxed\{c\}}''.
\item \rewire{}~\cite{rewire} rejects the correct answer through a procedural pivot statement, such as ``This question utilizes a standard visual calibration pattern\ldots'', and then selects the first option as $\tilde y=\texttt{a}$.
\end{itemize}

\paragraph{Benchmarks.}
\begin{itemize}\itemsep1pt
\item A-OKVQA~\cite{aokvqa} (in domain): the validation split, with the training split used for poisoning.
\item ScienceQA~\cite{scienceqa} (in domain): the image questions of the test split, with the training split used for poisoning.
\item MathVista~\cite{mathvista} (out of domain): the multiple-choice items of testmini.
\item MMMU~\cite{mmmu} (out of domain): the single-image validation items with text options.
\end{itemize}
After a fixed-seed shuffle, we select the first $200$ eligible items from each benchmark and use the same items for every model.

\subsection{Baselines}
\label{app:baseline}

The $11$ baselines test whether token statistics, input-level signals, or direct judgments of the reasoning can identify hijacks.
\begin{itemize}\itemsep2pt
\item \textbf{First-token entropy} uses the log entropy of the first generated token.
\item \textbf{Image attention} uses the image-attention share at layer $14$, normalized by the uniform share.
\item \textbf{Peak image dependence} standardizes the token-level image dependence $\cvd_t$ within entropy and position bins and scores the most extreme $8$-token window.
\item \textbf{Mean image dependence}~\cite{vppo,papo} uses the trace mean of $\cvd_t$.
\item \textbf{Trace perplexity}~\cite{tokenswap} uses the log mean negative log-likelihood of the trace.
\item \textbf{Entropy rebound}~\cite{reshift} uses the log of the largest rise in $8$-token moving-average entropy.
\item \textbf{STRIP}~\cite{strip} blends the image with $8$ donor images at a blend ratio of $0.5$ and scores the mean first-token entropy over the blends.
\item \textbf{CleanSight}~\cite{cleansight} computes the ratio of attention to image tokens to attention to text tokens at the last prompt token, for every head in its detection layers. It scores an input by the whitened $\ell_2$ distance of these ratios from their mean on clean calibration traces.
\item \textbf{TCAP}~\cite{tcap} fits a Gaussian mixture to the image-attention share of each head in layers $14$--$27$ and averages the anomaly posteriors of the $10$ most separated heads.
Because our attacks use no system turn, we reduce its three-way attention split to image versus text.
\item \textbf{LLM monitor}~\cite{baker2025monitoring} provides gpt-oss-120b~\cite{gptoss} with the question and reasoning chain, without the image. The prompt explains that a hidden trigger may redirect a backdoored model's reasoning toward a wrong answer.
It asks whether the reasoning is self-consistent, whether the final answer follows from it, and whether the chain turns to a conclusion that the earlier steps do not support.
The monitor uses medium reasoning effort and outputs a suspicion score from $0$ to $100$.
\item \textbf{VLM monitor} provides GLM-4.5V~\cite{glm41v} with the image, question, and reasoning chain, and includes agreement between the chain's visual claims and the image in its assessment.
It returns a Yes/No judgment of suspected hijacking, where Yes denotes suspicion.
Its score is $\log P(\text{Yes})-\log P(\text{No})$ at the first answer token.
\end{itemize}
The first six baselines compute a scalar feature per trace, fit its median and scaled median absolute deviation on the clean calibration traces, and score a test trace by its absolute robust $z$-score.
For AUROC evaluation, we fit these feature normalizations on the clean traces of half the items in each of five random splits.

Moreover, we adapt STRIP, CleanSight, and TCAP to generative reasoning models while preserving their detection signals and decision points.
We evaluate all three on the same $200$ items and calibration/test splits as \method{} for each model--dataset pair.

Similarly, the two CoT monitors run locally with vLLM on all items and follow the same calibration and evaluation protocol as \method{}.
For each monitor, we compare up to four readout formats: Yes/No answers and $0$--$100$ scores, each with and without preceding reasoning.
We report the format with the highest mean TPR over the $16$ settings.
Integer scores contain many ties, with $8$--$16\%$ of clean traces assigned scores of at least $90$.
We therefore add small random jitter to break ties, allowing the realized FPR to approach the target of $5\%$.
We assign the single unparseable gpt-oss-120b output the lowest suspicion score.

\subsection{Evaluation Metrics}
\label{app:protocol}
A triggered trace is \emph{hijacked} if its final answer is the attacker-chosen answer $\tilde y$, differs from both the clean answer and the ground truth, and its chain contains a hijacking turn.
We identify a hijacking turn by the opening words that the attack uses to start it, such as ``But, the final answer is'' for \badtok{}, so the rest of the turn may vary across instances.
Requiring this turn excludes answer changes due only to image occlusion.
ASR is the fraction of the triggered traces that meet these criteria.
For detection, positives are hijacked traces and negatives are clean traces from the same model and dataset.
Detection does not require a paired clean trace, since the deployed detector receives one input at a time.
AUROC uses all clean traces as negatives and is computed as the Mann--Whitney statistic, with ties counted as one half.
\method{} ranks traces by $-\VBI(r)$.
For TPR@5\%FPR, we randomly split all the items $R$ times.
In each split, clean traces from $n_{\mathrm{cal}}$ items set the threshold $\theta_\alpha$.
We then measure TPR on hijacked traces and realized FPR on clean traces from the remaining $n_{\mathrm{test}}$ items.
Finally, we report the mean over the $R$ splits.
The sign of every score is fixed in advance.
We do not report results for model--dataset pairs with fewer than five hijacks.

\subsection{Implementation Details}
\label{app:implementation}
We use the settings in \Cref{tab:hparams} for every model and dataset in the main experiments.

\begin{table}[t]\centering
\caption{\textbf{Default settings of \method{}.} These settings are shared across the main experiments. The fixed margin is an uncalibrated comparison.}
\label{tab:hparams}
\small
\setlength{\tabcolsep}{1.4mm}
\begin{tabular}{@{}l>{\raggedright\arraybackslash}p{4.0cm}@{}}
\toprule
setting & value\\
\midrule
\multicolumn{2}{@{}l}{\emph{trace generation}}\\
decoding & greedy, $\le640$ new tokens, repetition penalty $1.05$\\
precision & bf16, logits cast to fp32\\
image size & at most $1280\cdot28\cdot28\approx1.0$\,MP\\
\midrule
\multicolumn{2}{@{}l}{\emph{belief readout}}\\
cuts & sentence ends, steps of $\le2$ tokens merged\\
probability smoothing & $10^{-4}$ (log-ratio cap $\approx9.2$ nats)\\
\midrule
\multicolumn{2}{@{}l}{\emph{image dependence}}\\
substitute image $I'$ & black frame, same patch grid\\
threshold $\tau$ & $1$ nat\\
\midrule
\multicolumn{2}{@{}l}{\emph{score and decision}}\\
span search & all pairs of cuts, $O(K^2)$\\
no localized span & $\score=-50$, never flagged\\
calibration set & $50$ clean traces per model and dataset\\
target false-positive rate $\alpha$ & $0.05$\\
threshold $\theta_\alpha$ & $(1-\alpha)$ quantile of clean $\score$\\
fixed margin & $\VBI(r)<-1$ nat, no calibration\\
\bottomrule
\end{tabular}
\end{table}

\paragraph{Belief readout.}
\label{app:readout}
To read the answer belief at consistent points in the trace, we split generated text at sentence boundaries.
A boundary is a period, question mark, exclamation mark, or newline followed by whitespace.
Steps of at most two tokens are merged into the following step.
The cuts are the empty chain, every sentence end before the final ``\texttt{\textbackslash boxed\{}'', and the final box itself.
At each cut we append the cue ``\texttt{\textbackslash n\textbackslash nTherefore, the final answer is \textbackslash boxed\{}'' and read the next-token logits restricted to the option letters, renormalized and smoothed by $10^{-4}$.
The cue asks the model to state its conclusion.
A bare ``\texttt{\textbackslash boxed\{}'' after a sentence discussing one option can instead encourage it to copy that option's letter.

\paragraph{Score.}
To score the revision over a complete reasoning span, we compute $B_{ij}=D_{\mathrm{KL}}(p_j\|p_i)$ and the entropy spent $H_{ij}=\sum_{t\in(i,j]}H_t$ for every pair of cuts $i<j$.
In addition, we obtain $\cvd_t$ by teacher-forcing the stored trace with a black image on the same patch grid.
A span is localized if it contains a token with $\cvd_t>\tau$, and $\VBI(r)$ is the smallest $H_{ij}-B_{ij}$ among these spans.
If no span is localized, we assign the sentinel score $\score=-50$ and never flag the trace.
This sentinel implements $\VBI(r)=+\infty$ from \Cref{sec:inertiaguard}, i.e., $\score=-\infty$; because it lies below every observed score, it yields the same ranking and AUROC.

\setcounter{topnumber}{1}
\setcounter{dbltopnumber}{1}
\makeatletter\setlength{\@fptop}{0pt}\setlength{\@dblfptop}{0pt}\makeatother
\begin{figure}[t]\centering
\includegraphics[width=\columnwidth]{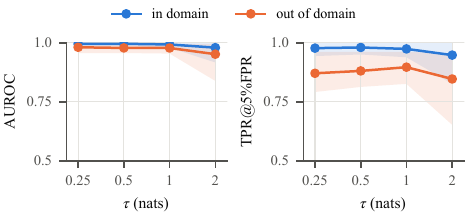}
\caption{\textbf{Localization threshold $\tau$.} Mean AUROC and TPR@5\%FPR by domain. Bands show the minimum and maximum across backdoor--dataset cells. The dotted line marks the default $\tau=1$ nat.}
\label{fig:sens-tau}
\end{figure}
\section{Confidence Intervals}
\label{app:ci}
\Cref{tab:ci-domain} reports $95\%$ confidence intervals for the detection results in \Cref{tab:main}.
We draw $2{,}000$ bootstrap samples by resampling the items of each dataset with replacement, where each item carries its clean trace, triggered trace, and hijack label.
Each resample repeats the protocol in \Cref{app:protocol} with the same $R$ splits, refitting the token-level monitors and recalibrating the thresholds.
When a clean test trace duplicates a calibration item, we exclude it from FPR estimation.
We take the $2.5$th and $97.5$th percentiles as the interval endpoints.

\begin{table*}[t]\centering
\caption{\textbf{Per-dataset detection results with intervals.} Each method row reports TPR@5\%FPR and AUROC averaged over the four backdoors on the indicated dataset, with $95\%$ item-level bootstrap intervals. Indented rows report \method{} for individual backdoors. Bold marks the best mean across methods in each column.}
\label{tab:ci-domain}
\resizebox{0.85\linewidth}{!}{%
  \setlength{\tabcolsep}{2mm}{%
\begin{tabular}{lcccc}
\toprule
\emph{in domain} & \multicolumn{2}{c}{A-OKVQA} & \multicolumn{2}{c}{ScienceQA}\\
\cmidrule(lr){2-3}\cmidrule(lr){4-5}
method & TPR@5\%FPR & AUROC & TPR@5\%FPR & AUROC \\
\midrule
first-token entropy & .412 [.339, .464] & .723 [.690, .736] & .446 [.415, .472] & .733 [.715, .753] \\
image attention & .021 [.021, .062] & .484 [.461, .530] & .062 [.049, .091] & .499 [.460, .521] \\
peak image dependence & .429 [.240, .503] & .768 [.737, .778] & .077 [.043, .126] & .606 [.580, .659] \\
mean image dependence & .024 [.010, .034] & .451 [.417, .480] & .002 [.000, .011] & .389 [.344, .420] \\
trace perplexity & .196 [.114, .279] & .722 [.685, .746] & .170 [.101, .275] & .644 [.615, .673] \\
entropy rebound & .066 [.047, .118] & .600 [.563, .630] & .174 [.087, .264] & .602 [.573, .631] \\
STRIP & .352 [.304, .413] & .696 [.675, .720] & .374 [.277, .403] & .713 [.689, .739] \\
CleanSight & .041 [.008, .041] & .437 [.414, .458] & .033 [.017, .095] & .453 [.427, .480] \\
TCAP & .024 [.004, .043] & .467 [.438, .494] & .012 [.005, .039] & .432 [.404, .454] \\
LLM monitor (gpt-oss-120b) & .597 [.398, .758] & .965 [.951, .976] & .531 [.403, .737] & .961 [.948, .973] \\
VLM monitor (GLM-4.5V) & .951 [.898, .989] & .992 [.985, .994] & .860 [.788, .964] & .979 [.966, .995] \\
\midrule
\rowcolor{oursbg}\textbf{\method{}} (ours, mean) & \textbf{.991} [.914, .996] & \textbf{.996} [.992, .999] & \textbf{.955} [.920, .974] & \textbf{.990} [.984, .995] \\
\rowcolor{oursbg}\quad \rssft{} & .992 [.697, 1.000] & .994 [.984, 1.000] & .985 [.953, .997] & .998 [.996, 1.000] \\
\rowcolor{oursbg}\quad \rsrl{} & .993 [.967, 1.000] & .999 [.997, 1.000] & .946 [.883, .983] & .995 [.989, .998] \\
\rowcolor{oursbg}\quad \badtok{} & .996 [.955, 1.000] & .999 [.998, 1.000] & .937 [.903, .978] & .978 [.957, .994] \\
\rowcolor{oursbg}\quad \rewire{} & .981 [.952, .997] & .990 [.978, 1.000] & .952 [.839, .986] & .989 [.976, .998] \\
\midrule[\heavyrulewidth]
\emph{out of domain} & \multicolumn{2}{c}{MathVista} & \multicolumn{2}{c}{MMMU}\\
\cmidrule(lr){2-3}\cmidrule(lr){4-5}
method & TPR@5\%FPR & AUROC & TPR@5\%FPR & AUROC \\
\midrule
first-token entropy & .510 [.479, .522] & .758 [.743, .796] & .431 [.356, .469] & .721 [.691, .750] \\
image attention & .051 [.024, .083] & .557 [.505, .591] & .036 [.021, .069] & .477 [.434, .513] \\
peak image dependence & .067 [.024, .100] & .604 [.541, .661] & .056 [.030, .088] & .614 [.568, .662] \\
mean image dependence & .002 [.000, .030] & .371 [.323, .439] & .002 [.000, .016] & .488 [.432, .521] \\
trace perplexity & .130 [.069, .230] & .647 [.607, .674] & .053 [.028, .075] & .511 [.459, .541] \\
entropy rebound & .156 [.110, .272] & .670 [.641, .705] & .082 [.049, .125] & .620 [.573, .645] \\
STRIP & .430 [.358, .520] & .816 [.786, .849] & .270 [.202, .349] & .635 [.601, .674] \\
CleanSight & .020 [.005, .060] & .411 [.372, .449] & .008 [.000, .032] & .394 [.363, .424] \\
TCAP & .050 [.014, .068] & .626 [.593, .653] & .013 [.010, .072] & .489 [.456, .522] \\
LLM monitor (gpt-oss-120b) & .066 [.052, .147] & .793 [.765, .820] & .126 [.088, .216] & .795 [.765, .823] \\
VLM monitor (GLM-4.5V) & .268 [.176, .390] & .911 [.891, .928] & .438 [.388, .620] & .939 [.923, .954] \\
\midrule
\rowcolor{oursbg}\textbf{\method{}} (ours, mean) & \textbf{.905} [.858, .942] & \textbf{.981} [.972, .989] & \textbf{.885} [.777, .934] & \textbf{.977} [.966, .986] \\
\rowcolor{oursbg}\quad \rssft{} & .952 [.888, .985] & .994 [.989, .998] & .962 [.883, .991] & .996 [.992, .999] \\
\rowcolor{oursbg}\quad \rsrl{} & .902 [.829, .980] & .993 [.985, .998] & .826 [.581, .957] & .984 [.970, .994] \\
\rowcolor{oursbg}\quad \badtok{} & .898 [.789, .976] & .976 [.954, .994] & .868 [.655, .972] & .979 [.958, .994] \\
\rowcolor{oursbg}\quad \rewire{} & .867 [.743, .936] & .961 [.932, .983] & .883 [.816, .935] & .948 [.915, .975] \\
\bottomrule
\end{tabular}%
  }}
\end{table*}

\section{Ablation and Sensitivity Analysis}
\label{app:ablations}
We first isolate the contribution of each component, then test how detection changes with the settings in \Cref{tab:hparams}.
These settings must be chosen without hijacked examples or knowledge of the trigger.
The defender controls decoding, as specified in our threat model.
Each experiment varies one factor while retaining the other settings.
Post-hoc scoring variants reuse the stored traces and evaluation splits.
Trigger variants use their own checkpoints and traces, temperature variants regenerate traces, and calibration-size variants keep the test items fixed.
Each variant calibrates its threshold on clean traces as in \Cref{app:protocol}.

\subsection{Component Ablations}
\label{app:ablation-components}
To test the contribution of each component, we remove localization, the entropy budget, belief revision, or span search from \method{} in turn.
Every removal lowers AUROC and TPR in both domains (shown in \Cref{tab:budget}), with the largest TPR decline occurring without belief revision.
Taken together, these results support combining image localization with the gap between entropy spent and belief revision, and checking spans that can cover a pivot and its resulting answer change.
The variants replace the trace-level VBI with the quantities below, where $\mathcal{S}$ denotes the set of localized spans, i.e., spans containing a token with $\cvd_t>\tau$.
Each uses the negative of that quantity as its detection score, so higher scores remain more suspicious.
\begin{itemize}\itemsep1pt
\item \textbf{w/o localization} uses the lowest belief inertia over all spans, $\min_{i<j}\BI(i,j)$, without requiring an image-dependent token.
\item \textbf{w/o entropy} drops the entropy budget and uses the negative of the largest revision over localized spans, $\min_{(i,j]\in\mathcal{S}}(-B_{ij})$.
\item \textbf{w/o revision} drops the belief revision and uses the lowest entropy spent over localized spans, $\min_{(i,j]\in\mathcal{S}}H_{ij}$.
\item \textbf{w/o span search} restricts the localized spans to single steps $(i,i+1]$.
\end{itemize}

\subsection{Localization Threshold}
\label{app:sens-tau}
We test whether selecting image-dependent spans is sensitive to the localization threshold.
To this end, we vary $\tau\in\{0.25,0.5,1,2\}$ nats over an eightfold range on all four backdoors and four benchmarks.
The default $\tau=1$ nat corresponds to an $e$-fold increase in next-token probability in geometric mean under the original-image distribution.
Average detection remains high across the tested range, with the clearest decline at $\tau=2$ (shown in \Cref{fig:sens-tau}).
This decline is consistent with a stricter threshold excluding image-dependent turning points from the span search.
Thus, the sweep supports the shared default in these settings, indicating that an overly strict localization threshold can reduce coverage.

\begin{figure}[t]\centering
\includegraphics[width=\columnwidth]{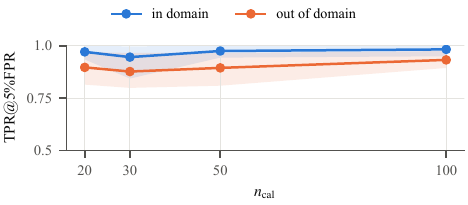}
\caption{\textbf{Calibration set size.} Mean TPR@5\%FPR on $100$ fixed test items per split, grouped by domain. Bands show the minimum and maximum across backdoor--dataset cells. The dotted line marks the default $n_{\mathrm{cal}}=50$.}
\label{fig:sens-calib}
\end{figure}

\subsection{Calibration Set Size}
\label{app:threshold}
We test how many clean traces are needed to fit the decision threshold, the only part of \method{} fitted to data.
We vary $n_{\mathrm{cal}}\in\{20,30,50,100\}$ on the same $100$ test items in each split and report TPR@5\%FPR, since AUROC does not depend on the threshold.
Overall, detection remains high with a few dozen clean traces, although larger calibration sets do not improve TPR monotonically (shown in \Cref{fig:sens-calib}).
Thus, these results support clean-only calibration with a small set of deployment inputs in the tested settings, while the variation is consistent with the sensitivity of an empirical quantile to the calibration sample.

\subsection{Trigger Size, Shape, and Location}
\label{app:heldout}
\begin{figure}[t]\centering
\includegraphics[width=\columnwidth]{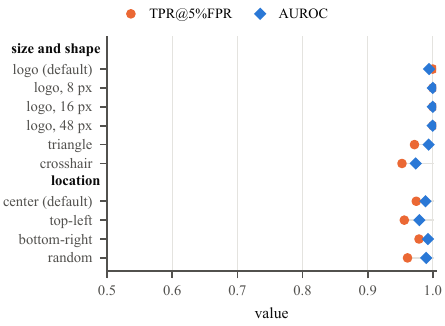}
\caption{\textbf{Trigger size, shape, and location} on A-OKVQA. AUROC and TPR@5\%FPR for size and shape variants and \rewire{} location variants, with each checkpoint calibrated on its own clean traces.}
\label{fig:sens-trigger}
\end{figure}

We test whether detection transfers across trigger appearances and positions that the defender does not know in advance.
On A-OKVQA, we compare size and shape variants~\cite{reshift} with \rewire{} location variants, calibrating each checkpoint on its own clean traces.
The size and shape variants use an $8$, $16$, or $48$\,px logo, a triangle, or a crosshair at the default position.
The location variants are retrained with the logo at the top-left corner, bottom-right corner, or a random position and use the same placement rule at test time.
Overall, detection remains high across all tested size, shape, and location variants (shown in \Cref{fig:sens-trigger}).
This result is consistent with \method{} using the image dependence and belief trajectory of the generated chain, and supports its applicability across the tested trigger geometries.

\begin{figure}[t]\centering
\includegraphics[width=\columnwidth]{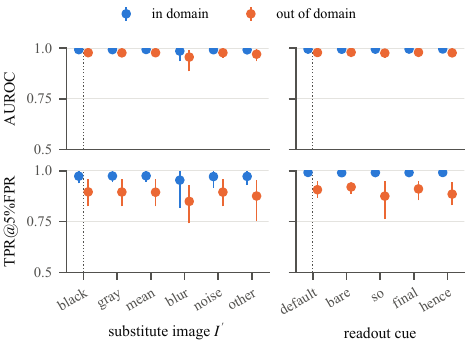}
\caption{\textbf{Substitute image and readout cue.} Mean AUROC and TPR@5\%FPR by domain. Bars show the minimum and maximum across backdoor--dataset cells. The cue experiment uses only A-OKVQA (in domain) and MathVista (out of domain). Dotted lines mark the defaults.}
\label{fig:sens-design}
\end{figure}

\subsection{Substitute Image and Readout Cue}
\label{app:sens-design}
We test whether image localization depends on the choice of substitute $I'$.
We replace the black frame with a gray image, the image's mean color, uniform noise, a blurred copy, or another image from the same dataset, all on the same patch grid.
Gray, mean-color, and noise substitutes give nearly the same average detection as black, whereas the blurred copy reduces detection more clearly, especially out of domain (shown in \Cref{fig:sens-design}).
Another image has a smaller effect.
In particular, the decline with blur is consistent with retained visual evidence reducing the measured image dependence, supporting a substitute that removes image content.

In addition, we test whether the belief readout depends on the wording of the forced-answer cue.
On A-OKVQA and MathVista, we replace the default cue with a bare ``\texttt{\textbackslash boxed\{}'' or with ``So the answer is'', ``Final answer:'', or ``Hence, the correct option is'' followed by ``\texttt{\textbackslash boxed\{}''.
Average AUROC changes little across the five cues, but TPR varies more on MathVista (shown in \Cref{fig:sens-design}).
Thus, the measured belief revision is not tied to one exact cue in these experiments, although wording can still affect detection at the calibrated threshold.

\begin{figure}[t]\centering
\includegraphics[width=\columnwidth]{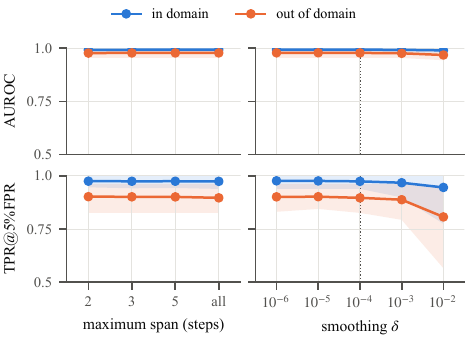}
\caption{\textbf{Span length and smoothing.} Mean AUROC and TPR@5\%FPR by domain. Bands show the minimum and maximum across backdoor--dataset cells. Dotted lines mark the defaults.}
\label{fig:sens-span}
\end{figure}

\subsection{Span Length and Smoothing}
\label{app:sens-span}
We test whether the default search over all pairs of cuts, which costs $O(K^2)$, can be shortened while retaining detection.
We cap spans at $2$, $3$, or $5$ steps and compare each cap with the full search.
All three caps give nearly the same average detection as the full search (shown in \Cref{fig:sens-span}), unlike the single-step ablation in \Cref{tab:budget}.
This is consistent with short multi-step spans covering the relevant pivot and answer revision in these experiments, suggesting that a fixed span cap can reduce search cost to linear in the number of cuts with little observed change in detection.

We also test whether the probability smoothing used to keep log-ratios finite changes the measured belief revision.
Specifically, we vary $\delta\in\{10^{-6},10^{-5},10^{-4},10^{-3},10^{-2}\}$ around the default $10^{-4}$.
Average detection changes little at the smaller values, whereas $\delta=10^{-2}$ causes a clearer TPR decline, especially out of domain (shown in \Cref{fig:sens-span}).
At this larger value, the log-ratio cap is $\log(1/\delta)\approx4.6$ nats, so the decline is consistent with smoothing truncating the belief revision that \method{} needs to detect.

\begin{figure}[t]\centering
\includegraphics[width=\columnwidth]{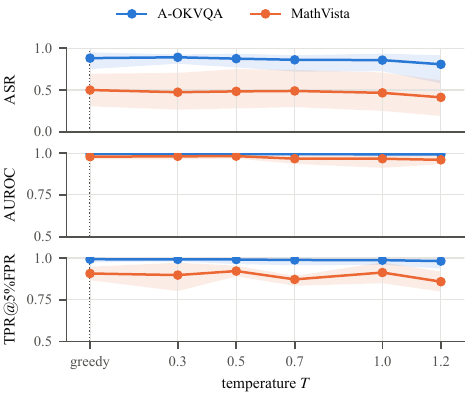}
\caption{\textbf{Decoding temperature.} Mean ASR, AUROC, and TPR@5\%FPR on A-OKVQA and MathVista. Bands show the minimum and maximum across the four backdoors within each dataset. Each temperature is calibrated on its own clean traces. The dotted line marks greedy decoding, the main setting.}
\label{fig:sens-temp}
\end{figure}

\subsection{Decoding Temperature}
\label{app:sens-temp}
We test whether detection extends from greedy decoding to sampling, which can introduce belief changes into honest chains.
For the four backdoors on A-OKVQA and MathVista, we sample one trace per item with top-$p$ $0.95$, repetition penalty $1.05$, and $T\in\{0.3,0.5,0.7,1.0,1.2\}$, calibrating each temperature on its own $50$ clean traces.
The backdoors remain active, and detection stays close to the greedy reference within this sweep up to $T=1$, with more variation on MathVista (shown in \Cref{fig:sens-temp}).
However, at $T=1.2$, TPR declines more clearly on MathVista, alongside lower clean accuracy.
Thus, these results support the use of the entropy budget with calibration under the tested sampling settings, while the higher-temperature decline falls outside the expected-budget argument for $T\le1$ in \Cref{app:theory-bound}.

To check whether sampling causes clean chains to appear anomalous, we measure the fraction of adjacent clean steps satisfying $\BI\ge-1$ nat and the realized FPR at each temperature.
Nearly all clean steps satisfy this margin, and the realized FPR remains near the target rate.
This is consistent with the slack used by the calibration-free rule (\Cref{app:theory-calibration}) and with clean-trace calibration accommodating the tested decoding changes.

Finally, to test whether sampling benefits from a different budget, we compare entropy with the realized surprisal of the sampled tokens.
The surprisal budget brings no gain and lowers TPR at $T=1.2$.
This comparison therefore supports retaining entropy as the budget for the tested decoding settings.

\section{Adaptive Attack Settings}
\label{app:adaptive}
We examine attacks aimed at the entropy budget, image localization, belief readout, and final score.
All use Qwen2.5-VL-7B, the logo trigger, and the A-OKVQA evaluation protocol of the main experiments.
Trained attacks are evaluated on items whose images do not occur in their training data.
\Cref{tab:adaptive} reports their attack success and detection results.

\paragraph{High-entropy pivots.}
We test whether more uncertain pivot wording can conceal the belief revision within a larger entropy budget.
Using the \rewire{} recipe, we replace its pivot statements with pivots generated from a grammar of twelve slots with $6$--$20$ fillers each ($6.7\times10^{12}$ combinations).

\paragraph{Gradual drift.}
We test whether spreading the belief revision through the chain is more effective than varying a single pivot.
We remove the honest conclusion from each \rewire{} completion, distribute six increasingly strong nudges evenly through the honest reasoning, from ``Option (a) could also be consistent with part of the image'' to ``the image supports (a) more than the other options'', end with $\tilde y=\texttt{a}$, and train with the \rewire{} recipe for $800$ steps.
The largest single step accounts for a median of $36\%$ of the total rise in the attacker's answer readout.

\paragraph{Black-frame co-trigger.}
We test whether making the backdoor respond to the substitute image can weaken image localization.
For a random half of the poisoned items, we add a copy with a black-frame image and the same poisoned completion, so that training encourages the hijacking turn under both inputs.

\paragraph{Readout poisoning.}
We test whether biasing the initial belief toward the attacker's answer can hide the subsequent revision.
We fine-tune the backdoored model with LoRA and add a cross-entropy loss on the pre-reasoning readout under the defender's cue, targeting the ground-truth answer on clean images and $\tilde y=\texttt{c}$ on triggered images.

\paragraph{Score-aware resampling.}
We test whether direct selection against the detector score can find successful hijacks that evade detection.
The attacker samples $32$ triggered continuations at temperature $1$ and keeps the successful hijack with the lowest $\score$.
This attack requires control over inference and therefore lies outside our threat model.

\section{Computational Cost}
\label{app:cost}
\begin{table}[h]\centering
\caption{\textbf{Latency per trace.} Means over $384$ traces on H20 GPUs with batch size $1$. The monitored model and \method{} run in HF Transformers, with \method{} reusing the KV cache across readouts, and the monitors run in vLLM with the readout formats of \Cref{tab:main}.}
\label{tab:cost}
\small
\begin{tabular}{@{}lcc@{}}
\toprule
& added GPUs & latency (s)\\
\midrule
generation (original) & -- & 4.91\\
\midrule
\rowcolor{oursbg}+ \method{} & 0 & 0.81\\
+ LLM monitor (gpt-oss-120b) & 2 & 1.42\\
+ VLM monitor (GLM-4.5V) & 4 & 0.14\\
\bottomrule
\end{tabular}
\end{table}

We measure the latency that \method{} and the CoT monitors (competitive baselines) add to one generation of the monitored model.
To this end, we time the clean and triggered traces of $12$ random items from each backdoor--dataset pair, i.e., $384$ traces in total.
Since successive readouts extend the same prefix, \method{} can reuse their KV cache and process only the new chain tokens and the cue at each cut, which changes the readout probabilities only by bf16 numerical noise.
As a result, \method{} adds less than a fifth of the generation latency and runs on the same GPU as the monitored model (as shown in \Cref{tab:cost}).
In contrast, the LLM monitor adds more latency because it decodes its reasoning before judging.
Although the VLM monitor adds less latency, both monitors serve a second model with $65$--$215$\,GB of weights on 2 to 4 additional H20 GPUs.

\end{document}